\documentclass{article}
\usepackage{spconf,amsmath,graphicx,hyperref}

\title{Disentangling Foreground and Background for vision-Language Navigation via Online Augmentation}
\name{Yunbo Xu$^*$, Xuesong Zhang$^*$, Jia Li$^\dagger$ \thanks{$^*$ denotes equal contribution, and $^{\dagger}$ indicates the corresponding author.}, Zhenzhen Hu, Richang Hong}

\address{Hefei University of Technology, Hefei, China}
\usepackage{caption}
\usepackage{multirow}
\usepackage{xcolor}
\usepackage{amsmath,amssymb,amsfonts}
\usepackage{algorithmic}
\usepackage{graphicx}
\usepackage{colortbl}

\usepackage{booktabs}

\usepackage{textcomp}

\begin{document}
\ninept

\maketitle
\begin{abstract}

Following language instructions, vision-language navigation (VLN) agents are tasked with navigating unseen environments. 
While augmenting multifaceted visual representations has propelled advancements in VLN, the significance of foreground and background in visual observations remains underexplored. Intuitively, foreground regions provide semantic cues, whereas the background encompasses spatial connectivity information. 
Inspired on this insight, we propose a \textbf{C}onsensus-driven \textbf{O}nline \textbf{F}eature \textbf{A}ugmentation strategy (\textit{COFA}) with alternative foreground and background features to facilitate the navigable generalization.
Specifically, we first leverage semantically-enhanced landmark identification to disentangle foreground and background as candidate augmented features. 
Subsequently, a consensus-driven online augmentation strategy encourages the agent to consolidate two-stage voting results on feature preferences according to diverse instructions and navigational locations.
Experiments on REVERIE and R2R demonstrate that our online foreground-background augmentation boosts the generalization of baseline and attains state-of-the-art performance.

\end{abstract}
\section{Introduction}
\label{sec:intro}







Vision-and-Language Navigation (VLN) aims to develop an egocentric agent capable of following natural language instructions to navigate through previously unseen environments. Given its potential in real-world applications such as disaster rescue and assistive navigation for the visually impaired, VLN has attracted considerable research attention. Among existing benchmarks, \textit{R2R} \cite{anderson2018R2R} focuses purely on instruction-following navigation, while \textit{REVERIE} \cite{qi2020reverie} introduces the additional challenge of grounding and recognizing target objects described in instructions. Despite their promise, building well-trained VLN agents remains non-trivial, as agents must generalize to unseen environmental layouts by perceiving diverse visual observations. 


To complete the challenging navigation task, 
an promising direction involves applying data augmentation strategies to effectively enlarge the scale and diversity of training environments. For example, large-scale generation of photo-realistic environments \cite{wang2023scalevln} has been shown to substantially improve model performance, while FDA \cite{he2024fda} shifts the focus from spatial augmentations to frequency-based perturbations, thereby facilitating cross-environment generalization. 
More recent research has explored enhancing generalization by constructing diverse environmental representations, such as grid-based layouts \cite{wang2023gridmm}, topological maps \cite{chen2022duet}.
Although these methods have advanced VLN research, they simultaneously increase training costs due to the growing model parameters and training data. 

Additionally, many methods introduce external knowledge or additional modalities such as depth \cite{zhang2025agentjourneyrgbunveiling, tan2022depth} to complement RGB images in navigation decision-making. However, the intrinsic information within RGB images (e.g., foreground and background) has not been thoroughly explored or utilized. Neuroimaging studies \cite{papale2018foreground} have demonstrated that, during natural image viewing, visual cortical regions exhibit “foreground enhancement” and “background suppression” mechanisms, suggesting that foreground elements are more prominently encoded in neural representations. Yet, such insights may not consistently align with navigation demands. For instance, when following the instruction \textit{“walk through the corridor to the kitchen and find a mug”}, spatial layout from background regions suffices during the corridor traversal, whereas foreground objects (e.g., the mug) become critical upon entering the kitchen. Despite this intuitive and biologically inspired perspective, the role of foreground and background remains largely underexplored in VLN research.

\begin{figure*}[ht!]
    \centering
    \includegraphics[width=1\linewidth]{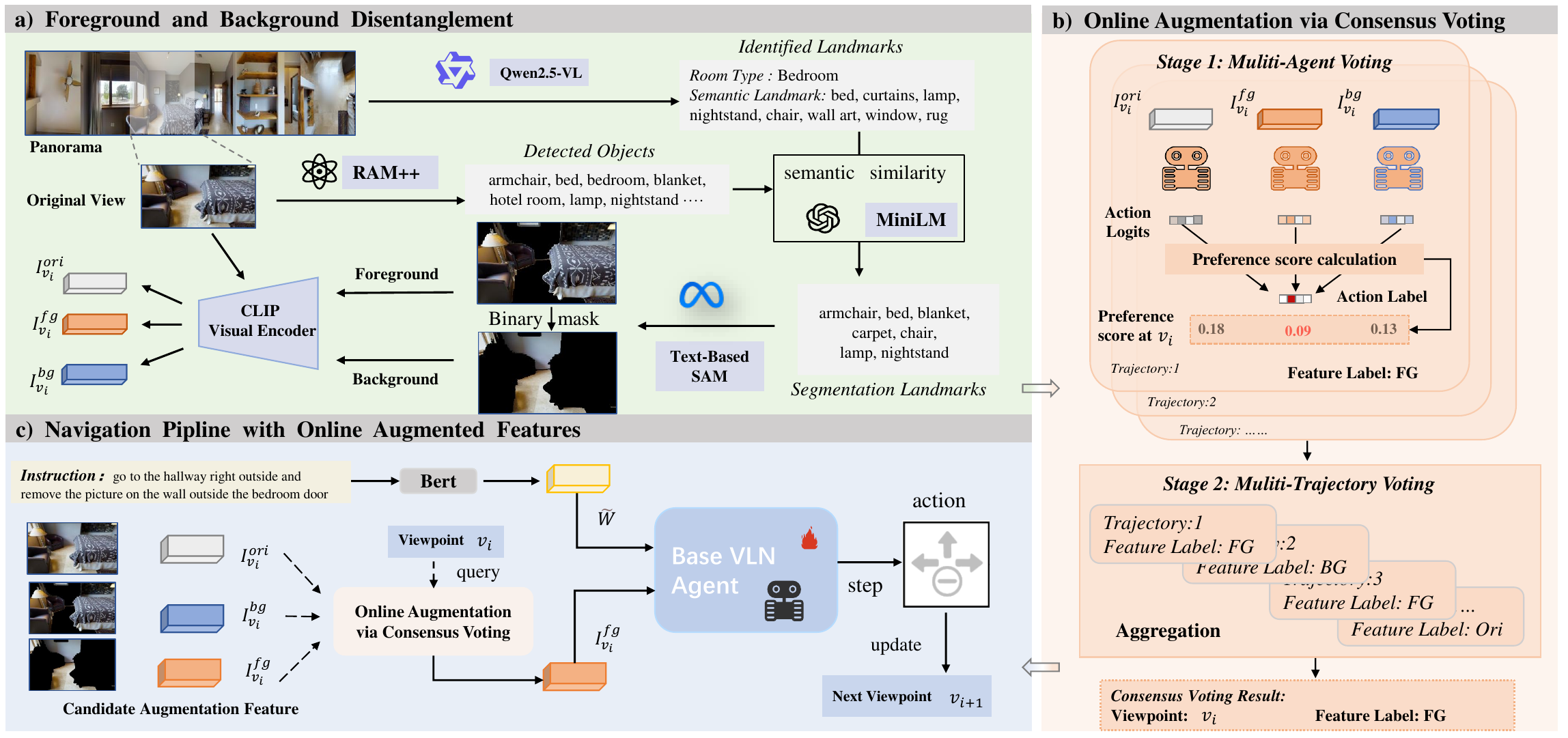}
    \caption{The overview of the proposed COFA: a) we extract foreground and background features by identifying spatially disentangled regions through foreground landmark identification; b)  online augmentation at the viewpoint level using two-stage voting for preferred augmentation features; c) the proposed online augmented features can be seamlessly integrated into a generic navigation pipeline.}
    \label{fig:cofa}
\vspace*{-0.2cm}
\end{figure*}

In this work, we propose a \textbf{C}onsenus-driven \textbf{O}nline \textbf{F}eature \textbf{A}ugmentation strategy (\textit{COFA}) as shown in Fig.\ref{fig:cofa}, which leverages spatially disentangled foreground and background features to address the aforementioned two challenges. 
We first semantically identify foreground landmarks and extract spatially disentangled foreground and background features.
Foreground objects are detected by an object detector, refined through landmark identification with a Qwen2.5-VL \cite{Qwen2.5-VL} and all-MiniLM-L6-v2 \cite{wang2020minilm}, and then separated into foreground and background regions using a text-driven segmentation model EVF-SAM \cite{zhang2024evf}.  Next, a CLIP visual encoder\cite{radford2021clip} further obtains the corresponding foreground and background features of each viewpoint. 
Subsequently, we employ an online feature augmentation mechanism that consolidates the agent’s viewpoint-level preferences from candidate features through a two-stage voting process. We further apply the consensus preference for feature augmentation to enrich environmental diversity, which in turn to enhance the navigational generalization of agents.
Unlike conventional offline augmentation methods \cite{li2022envedit, li2023panogen}, COFA achieves such enhancement with negligible training overhead without architectural modifications. 
Experiments on R2R and REVERIE demonstrate the superiority of our method, surpassing prior state-of-the-art and offline augmentation approaches. Additionally, we provide a quantitative analysis of the preference feature distribution across different datasets and splits.
Our key contributions are threefold:
\begin{itemize}
    \item We systematically identify and disentangle foreground and background information within visual environments to enhance and exploit the intrinsic diversity of images for VLN.
    \item We propose a novel online augmentation strategy that employs a two-stage voting mechanism to identify the preferred feature for each viewpoint with negligible additional cost.
    \item Extensive experiments on R2R and REVERIE convincingly demonstrate the effectiveness of augmented features, with the proposed COFA achieving state-of-the-art performance.
\end{itemize}

\section{Method}

\subsection{Overview of the Proposed Method}

Our proposed framework for discrete VLN tasks is illustrated in Fig.~\ref{fig:cofa}. We first disentangle foreground and background features as described in Sec.\ref{sec:disentagnle}. Based on these disentangled features, we then perform online feature augmentation, as detailed in Sec. \ref{sec:vote}, allowing the agent to dynamically adapt to diverse environments without relying on additional offline data generation. Finally, following a general VLN paradigm, the agent selects the preferred augmented feature at each viewpoint $v_i$ and combines it with the instruction $W$ to predict the next action, continuing this navigation process until reaching the target location or exceeding the step limit.

 
\subsection{Foreground and Background Disentanglement}
\label{sec:disentagnle}

To fully investigate the influence of foreground and background features within environmental observations during navigation episode, we designed a pipeline that integrates various off-the-shelf  VLMs to semantically identify foreground landmarks. This pipeline further spatially disentangles the foreground and background regions of the navigatin environment and extracts their respective visual features. As shown in Fig.\ref{fig:cofa} (a), it consists of two main components: \textit{Semantic-Enhanced Landmark Identification} and \textit{Spatially-Disentangled Feature Extraction}.

\textbf{\textit{Semantic-Enhanced Foreground Landmark Identification}:}
To achieve disentanglement between foreground and background regions, we first localize potential foreground objects using the object detection model RAM++ \cite{huang2023open}. However, RAM++ tends to detect all objects in the environment, including those belong to the background. To mitigate this, we leverage the semantic reasoning capability of VLM to filter objects according to their semantic relevance to the room type. Specifically, for each panoramic image at $v_i$, we provide Qwen2.5-VL \cite{Qwen2.5-VL} with a prompt: \textit{“Identify the room type and list 7–8 essential objects commonly found in this space, formatted as: Room Type: [type]; Key Objects: [object1, …, objectN]”}. Qwen2.5-VL then generates a set of semantically relevant landmarks. Finally, we compute the semantic similarity between all detected objects and the identified landmarks using the lightweight model all-MiniLM-L6-v2 \cite{wang2020minilm}, thereby filtering out irrelevant objects and retaining only the key landmarks for each view.

\textbf{\textit{Spatially-Disentangled Feature Extraction}}:
For each viewpoint $v_i$, we obtain 36 views, denoted as ${O_{v_{i}}=\{O_{v_{i}}^{1},O_{v_{i}}^{2},\ldots,O_{v_{i}}^{36}\}}.$ Given the landmark tags extracted for $v_i$, we use the text-driven segmentation model EVF-SAM \cite{zhang2024evf} to generate a binary mask $\mathbf{M}_{v_i}^j$ for each view $O_{v_i}^j$. The mask highlights the landmark-related regions in the image. We then perform element-wise multiplication to overlay the mask on the original image:
\begin{equation}
    \label{mask}
    \tilde{O}_{{v_{i}}}^{{j}}={O}_{{v_{i}}}^{{j}}\odot\mathbf{M}_{{v_{i}}}^{{j}},\quad{j}={1,\ldots,36},
\end{equation}
, where $\odot$ denotes pixel-wise multiplication. After obtaining the masked results for all 36 views, we stack them to form the disentangled representation $\tilde{O}_{v_i}^{fg}$. Finally, we feed $\tilde{O}_{v_i}^{fg}$ into CLIP-ViT-B/16 to extract the foreground feature $\mathbf{I}_{v_i}^{fg}$.
Similarly, the background feature $\mathbf{I}_{v_i}^{bg}$ is obtained by applying the complementary mask regions in Eq.\ref{mask}, following the same procedure as for the foreground.

\begin{table*}[ht!]
\centering
\caption{Comparison with the state of the art on REVERIE dataset. \textbf{Bold} and \underline{underlines} highlight the best and runner-up performance in each column, while the \colorbox{gray!20}{gray} row underscores our method. $\uparrow$ indicates better performance with higher values. $^\ddagger$ indicates that the method is based on offline augmentation.}
\begin{tabular}{l|ccccc|ccccc}
\toprule
\multirow{2}{*}{\textbf{Methods}}  &  \multicolumn{5}{c|}{\textbf{\textit{Val-unseen}}}    &
\multicolumn{5}{c}{\textbf{\textit{Test-unseen}}}\\& 
\multicolumn{1}{c}{TL} &
\multicolumn{1}{c}{SR$\uparrow$} & 
\multicolumn{1}{c}{SPL$\uparrow$}&
\multicolumn{1}{c}{RGS$\uparrow$} & 
\multicolumn{1}{c|}{RGSPL $\uparrow$} &TL  & SR $\uparrow$                & SPL$\uparrow$                & RGS$\uparrow$               & RGSPL$\uparrow$          \\      \hline
DSRG  \cite{wang2023dsrg}&-  & 47.83  & 34.02 & 32.69    & 23.37    & -   & 54.04  & 37.09   & 32.49   & 22.18         \\
BEVBert  \cite{an2023bevbert}  & -     & 51.78  &\underline{36.47}  & 34.71     & 24.44 & -   & 52.81   & 36.41  & 32.06  & 22.09       \\
GridMM  \cite{wang2023gridmm}   & 23.20  & 51.37    & \underline{36.47} & 34.57 & 24.56   &19.97& 53.13   & 36.60   & 34.87  & 23.45                \\
DAP \cite{liu2024dap}   &16.32     & 32.17               & 27.30                & 20.44                & 17.32   &   15.37             & 30.26              & 24.07                & 17.08               & 14.78                 \\

GAR \cite{zhou2025exploring}    &22.10     & 48.72               & 34.53                & 32.65                & \textbf{25.87}&   19.36             & 53.17              & 37.87                & 33.26               & 22.31                 \\

ViTeC \cite{gao2025visual}  &24.07  & 50.18   & 35.06   & \underline{34.82}  & 24.23    &23.30   & \textbf{57.52}  & 38.09   & 34.09   & 22.81   \\
\midrule
FDA$^\ddagger$\cite{he2024fda}  &19.04  & 47.57   & 35.90 & 32.06  & 24.31    &17.30    & 49.62    & 36.45    & 30.34  & 22.08    \\
RAM$^\ddagger$\cite{wei2025unseen}  &25.44  & \underline{51.89}  & 35.00   & 34.31  & 23.20    &22.78   & \underline{57.44}  & \underline{41.41}   & \underline{36.05}   & \underline{25.77} \\ 
\midrule
Baseline \cite{chen2022duet}  &22.11     & 46.98  & 33.73   & 32.15   & 23.03   & 21.30      & 52.51    & 36.06   & 31.88     & 22.06  \\
\rowcolor{blue!5} COFA (Ours) &24.85   &\textbf{54.62}  &  \textbf{38.17}& \textbf{36.07}& \underline{25.01}    & 18.92&55.15&\textbf{41.62} &\textbf{36.09} &\textbf{26.80} \\ 
 \bottomrule
\end{tabular}
\vspace*{-0.2cm}
\label{tab:reverie-sota}
\end{table*}

\subsection{Online Feature Augmentation via Consensus Voting}
\label{sec:vote}

Building on the spatially disentangled foreground and background features, we propose an online feature augmentation framework via consensus voting as illustrated in Fig.\ref{fig:cofa} (b). 
Unlike prior offline works \cite{li2023panogen, li2022envedit} that pre-generate extensive synthetic data, our approach performs viewpoint-level augmentation based on two-stage consensus voting, which selectively replace the most suitable feature among spatially-disentangled foreground $\mathbf{I}_{v_i}^{fg}$, background $\mathbf{I}_{v_i}^{bg}$, or original $\mathbf{I}_{v_i}^{ori}$ feature.

\subsubsection{Stochastic Online Augmentation}
To illustrate the advantage of our method, we first introduce a stochastic augmentation strategy for comparison. Specifically, this strategy selects candidate augmented features $\mathbf{I}_{v_i}^{aug}$ ($\mathbf{I}_{v_i}^{fg}$ or $\mathbf{I}_{v_i}^{bg}$) at $v_i$ with probability $p \sim \mathcal{U}(0,1)$, formalized as:
\begin{equation}
\label{eq:random}
\mathbf{I}_{v_i}^{\text{rand}} =
\begin{cases}
\mathbf{I}_{v_i}^{\text{aug}}, & \text{if } p > 0.5, \\
\mathbf{I}_{v_i}^{\text{ori}}, & \text{if } p \le 0.5,
\end{cases}
\end{equation}
where $\mathbf{I}_{v_i}^{\text{rand}}$ denotes the randomly selected feature at viewpoint $v_i$. However, this  augmentation may degrade navigation performance because suboptimal features are often selected at certain viewpoints. 

\subsubsection{Consensus-driven Online Feature Augmentation}
Our consensus-driven online method addresses this issue by dynamically selecting the most appropriate feature at each viewpoint. This approach relies on a two-stage consensus voting mechanism: \textit{Multi-agent Voting} and \textit{Multi-trajectory Voting}.

\textbf{\textit{Multi-agent Voting.}} We employ three agents, each pre-trained exclusively on one type of feature—$\mathbf{I}_{v_i}^{ori}$, $\mathbf{I}_{v_i}^{fg}$, and $\mathbf{I}_{v_i}^{bg}$. Formally, let $A_f$ denote the agent pretrained on feature type $f \in {\text{ori}, \text{fg}, \text{bg}}$. These agents perform parameter-frozen exploration on the training split. For each viewpoint $v_i$ within a trajectory $T_j$, we compute a preference score $S(A_f, v_i, T_j)$, defined as the cross-entropy between the predicted action logits and the ground-truth labels. The voting decision is made by selecting the feature label corresponding to the lowest preference score:
\begin{equation}
\hat{f}(v_i, T_j) = \underset{f \in {\text{ori}, \text{fg}, \text{bg}}}{\arg\min} S(A_f, v_i, T_j),
\end{equation}
where $\hat{f}(v_i, T_j)$ denotes the voted feature label for $v_i$ in trajectory $T_j$.

\textbf{\textit{Multi-trajectory Voting.}} Since a viewpoint $v_i$ may appear in multiple trajectories, feature augmentation based on a single trajectory may introduce bias. To alleviate this, we aggregate predictions across all trajectories containing $v_i$ and adopt a majority voting strategy:
\begin{equation}
\hat{f}^{\text{final}}_{v_i} = \underset{f \in \{\text{ori}, \text{fg}, \text{bg}\}}{\arg\max} \sum_{T_j \in {T}_{v_i}} \mathbb{I}[\hat{f}(v_i, T_j) = f],
\end{equation}

Here,  $\hat{f}^{\text{final}}_{v_i}$ represents the final voted feature label for viewpoint $v_i$ based on majority voting across all trajectories $T_{v_i}$containing $v_i$. The indicator function $\mathbb{I}(\cdot)$ ensures only trajectories sharing $v_i$ contribute to the voting process.

\textbf{\textit{Viewpoint-level Feature Augmentation.}} Based on the consensus voted results $\hat{f}^{\text{final}}_{v_i}$, we construct the online augmented visual feature $I_{v_i}^{oa}$ by selecting from the three candidate representations:
\begin{equation}
    \mathbf{I}_{v_i}^{\mathsf{oa}} = 
    \begin{cases}
        \mathbf{I}_{v_i}^{\mathsf{ori}} & \text{if } \hat{f}^{\mathrm{final}}_{v_i} = \mathrm{ori}, \\
        \mathbf{I}_{v_i}^{\mathsf{fg}}  & \text{if } \hat{f}^{\mathrm{final}}_{v_i} = \mathrm{fg},  \\
        \mathbf{I}_{v_i}^{\mathsf{bg}}  & \text{if } \hat{f}^{\mathrm{final}}_{v_i} = \mathrm{bg}.
    \end{cases}
\end{equation}

Building on alternative foreground and background features, COFA can be seamlessly applied to prior discrete VLN agents without architectural modifications and with negligible additional training cost.
For simplicity, we adopt the popular navigation pipeline (based on DUET \cite{chen2022duet}), as illustrated in Fig.~\ref{fig:cofa}(c). The proposed COFA ensures that each viewpoint is represented by the most beneficial and reliable feature, explicitly guiding the agent to select and focus on appropriate visual regions according to varying instructions and navigational locations. This design mitigates randomness and enhances the overall robustness of navigation.



\section{Experiments}
\label{sec:print}

\subsection{Experiments Setting}
\label{sec:page}

We conduct comprehensive experiments on two VLN benchmarks: R2R \cite{anderson2018R2R} with fine-grained path instructions and the REVERIE \cite{qi2020reverie} dataset for object-oriented navigation tasks. On R2R, we mainly report standard metrics including Success Rate (SR), SR weighted by Path Length (SPL) and Navigation Error (NE). On REVERIE, we additionally evaluate object grounding performance using Remote Grounding Success (RGS) and RGS weightd by Path Legnth (RGSPL). All experiments are conducted on a single NVIDIA RTX 4090 GPU. The model is first pre-trained using three proxy tasks: Masked Language Modeling, Step Action Prediction, and Object Grounding, with a batch size of 32 for 100k iterations. Subsequently, the model is fine-tuned with a batch size of 12 for 20k iterations. We maintain the baseline DUET \cite{chen2022duet} architecture unchanged.

\begin{table}[t!]
\centering
\caption{Comparison with state of the art on the R2R dataset.}
\begin{tabular}{l|ccc|ccc}
\toprule
\multirow{2}{*}{\textbf{Methods}} &\multicolumn{3}{c|}{\textbf{\textit{Val-unseen}}} & \multicolumn{3}{c}{\textbf{\textit{Test-unseen}}} \\  & SR $\uparrow$  & SPL$\uparrow$  & NE$\downarrow$  & SR$\uparrow$  & SPL$\uparrow$  & NE$\downarrow$ \\ \hline
HAMT \cite{chen2021hamt}  & 66   & 61   & 3.29  &  65  &  60  &  3.93
\\
EnvEdit \cite{chen2021hamt}  & 69   & \textbf{64.4}   & 3.29  &  65  &  60  &  3.93 \\

DAP \cite{liu2024dap}    & 65   & 59   & 3.62  &  64  &  59  &  3.95 \\ 
DSRG \cite{wang2023dsrg}   & 73   & 62   &3.00      & 72  &  61  &  3.33 \\
\midrule
FDA$^\ddagger$ \cite{he2024fda}  & 72   & 64   & 3.41 &  69  &  62  &  3.41 \\
RAM$^\ddagger$ \cite{wei2025unseen} &73.7  &63.1  & 2.96 & 71&61 & 3.34 \\
\midrule
Baseline\cite{chen2022duet}   & 71 & 61 & 3.30 &  69  &  59  &  3.65 \\

\rowcolor{blue!5} COFA (Ours)  & \textbf{74.2} & 64.2 & \textbf{2.92} & \textbf{74.7}  &  \textbf{62.6}  &  \textbf{2.86} \\
\bottomrule
\end{tabular}
\label{tab:r2r-sota}
\end{table}

\subsection{Comparisons with State-of-the-Art Methods}
We compare our method with state-of-the-art approaches on two prominent VLN datasets, R2R and REVERIE. In Tab. \ref{tab:reverie-sota}, our method significantly outperforms baseline across multiple metrics. Particularly, it yields gains of $4.44\%$ in SPL and $1.98\%$ in RGSPL on the Val-unseen split. On the Test-unseen split, we see similar robust improvements, with SPL up by ($+$5.56\%) and RGSPL by ($+$4.74\%).  Overall, our method establishes new state-of-the-art performance on REVERIE Val-unseen split, with SPL and RGS reaching 38.17 \% and 36.07 \%, respectively.
For R2R (Tab. \ref{tab:r2r-sota}), although not object-oriented, our method still achieves consistent improvements over the baseline both on Val-unseen and Test-unseen split, 
as shown in Tab. \ref{tab:r2r-sota}. Compared to baseline, we achieve notable gains in both SR ($+$3.26\%) and SPL ($+$3.24\%) on Val-unseen split. 
Overall, the propose an online augmentation approach with disentangled foreground-background features is simple yet effective, surpassing both prior state-of-the-art and offline augmentation methods.

\begin{table}[t]
    \centering
    \caption{Ablation of different feature type on REVERIE.}
    \begin{tabular}{l|c|cccc}
        \toprule
        ID & {Features} & SR$\uparrow$ & SPL$\uparrow$ & RGS$\uparrow$ & RGSPL $\uparrow$\\
        
        \midrule
        1 & Ori (ViT) & 46.98 & 33.73& 32.15 & 23.03 \\
        2 & Ori (CLIP) & 48.14 & 34.97 & 33.37 & 24.34 \\
        3 & BG  & 53.14 & 35.81 &35.10 & 23.81 \\
        4 & FG  & \textbf{54.44} & \textbf{37.20} & \textbf{36.59} & \textbf{25.34} \\

        \bottomrule
    \end{tabular}
    \label{tab:features}
\end{table}

\begin{table}[t]
    \centering
    \caption{Experimental results with different augmentation strategy on REVERIE Val-Unseen Split.}
    \begin{tabular}{l|c|cccc}
        \toprule
        \textbf{Strategy}&Feature & SR$\uparrow$ & SPL$\uparrow$ & RGS$\uparrow$ & RGSPL $\uparrow$  \\
        \midrule
        Stochastic &BG& 49.05 & 36.04 & 32.26 & 23.75  \\
        Stochastic &FG & 53.20 & 36.68 &35.05 & 24.68  \\
        Replace &BG& 53.14 & 35.81 & 35.10 & 23.81  \\
        Replace &FG& 54.44 & 37.20& \textbf{36.59} & \textbf{25.34}  \\
        \midrule
        COFA &FG+BG &\textbf{54.62} & \textbf{38.17} & 36.07 & 25.01  \\
        \bottomrule
    \end{tabular}
    \label{tab:augmentation}
\end{table}

\vspace{-0.1cm}
\subsection{Ablation Study}
\label{sec:illust}

To verify the effectiveness of our proposed online augmentation strategy with foreground-background, we conduct ablation studies on the object-oriented REVERIE Val-unseen dataset. 

\textbf{\textit{Disentangled Features.}} 
We first evaluate the proposed spatially disentangled features by directly replacing. As shown in Tab.\ref{tab:features}, results show that both foreground and background features consistently improve navigation performance. To exclude the possibility that these gains originate from the CLIP encoder itself, we further extract features from the original images using the same encoder. Although this yields better navigation performance compared with the baseline, the improvements remain notably inferior to those achieved by our proposed disentangled features.

\textbf{\textit{Augmentation Strategy.}}
Next, we examine different feature augmentation strategies to assess the effectiveness of our proposed COFA in Tab.\ref{tab:augmentation}. The simplest strategy, direct replacement, substitutes original features with our disentangled features. The second strategy, stochastic augmentation, is defined in Eq.(\ref{eq:random}). Finally, our proposed COFA. Experimental results reveal that stochastic augmentation often degrades navigation and object grounding performance due to suboptimal feature selection at viewpoint level. In contrast, our method consistently outperforms stochastic augmentation by assigning more appropriate features to each viewpoint. Compared with direct replacement, COFA exhibits a trade-off: while it markedly improves navigation performance, it slightly reduces object grounding accuracy. This occurs because, at certain viewpoints, background features emphasizing spatial layout benefit navigation but weaken the agent’s ability to recognize foreground regions, thereby harming object grounding. Since navigation is the primary goal in VLN, we consider this trade-off acceptable.

\subsection{Qualitative Results}
\begin{figure}[t]
    \includegraphics[width=0.99\linewidth]{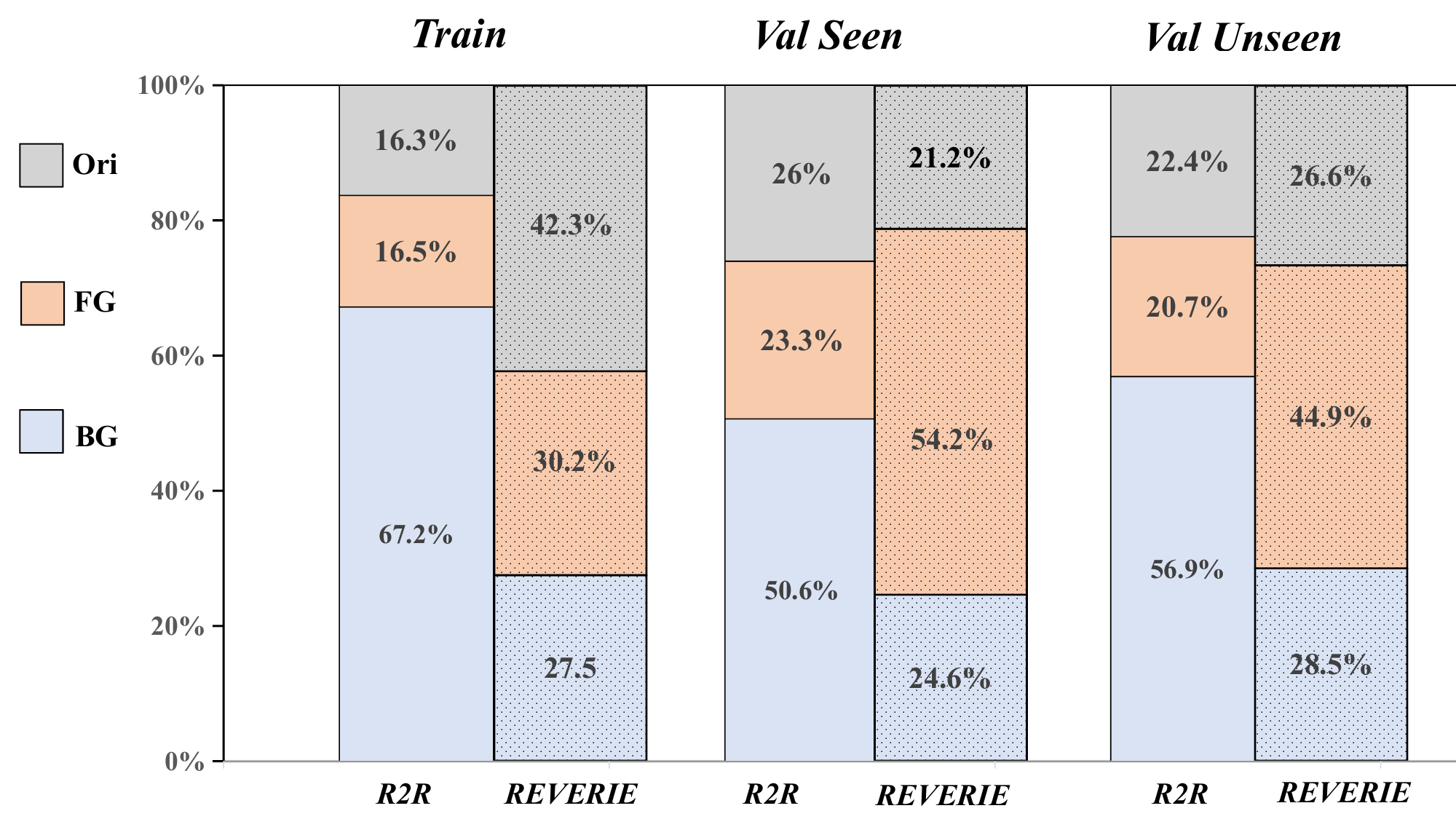}
    \caption{The quantitative analysis of viewpoint-level features preference across different VLN datasets and splits.}
    \label{fig:vis-2}
    \vspace{-10pt}
\end{figure}

As shown in Fig.~\ref{fig:vis-2}, we analyze viewpoint-level feature preferences across VLN datasets by aggregating consensus voting results. On the REVERIE~\cite{qi2020reverie} benchmark, where instructions emphasize salient landmarks but offer limited action guidance, agents prefer foreground features that enhance perception of object-relevant regions.
In contrast, on R2R\cite{anderson2018R2R}, where fine-grained action cues are explicit, agents rely more on background features, as spatial layout information better aligns visual observations with action instructions.

\section{Conclusion}
\label{sec:prior}

In this paper, we propose an online feature augmentation method, COFA, which leverages carefully disentangled foreground and background features to enhance environmental diversity. COFA employs a consensus-driven two-stage voting strategy to select appropriate features at each viewpoint, without introducing external environments or altering model architectures. Extensive experiments demonstrate that our method significantly boosts the generalization performance of baseline agents, achieving state-of-the-art results. Since our approach requires no additional environments and is agnostic to model architecture, it can be seamlessly and effectively extended to other VLN methods in future research.


\vfill\pagebreak

\clearpage

\newpage
\bibliographystyle{IEEEbib}
\bibliography{strings,arxiv}

\end{document}